\documentclass[sigconf,natbib=false,authorversion]{acmart}

\acmSubmissionID{248}
\usepackage{etoolbox}
\usepackage{xurl}

\usepackage{wrapfig}

\usepackage{latexsym,verbatim}
\usepackage{xspace}
\usepackage{xcolor}
\usepackage{graphicx}

\RequirePackage[
datamodel=acmdatamodel,
giveninits=true,
isbn=false,
url=false,
doi=false,
eprint=false,
mincrossrefs=1,
style=acmnumeric, % use style=acmauthoryear for publications that require it
]{biblatex}

\usepackage{multicol}
\usepackage[inline]{enumitem}
\setlist{nosep}

\newtheorem{intuition}{Intuition}

\newcommand{\ochoice}{\mathbf{Choice}}

\newcommand{\seq}[2][n]{#2_1,\dots,#2_{#1}}

\newcommand{\Obl}{\ensuremath{\mathsf{O}}\xspace}
\newcommand{\Perm}{\ensuremath{\mathsf{P}}\xspace}

\newcommand{\claim}{\ensuremath{\mathsf{Claim}}}
\newcommand{\dclaim}{\ensuremath{\mathsf{Claim}_{\OBL}}}
\newcommand{\fclaim}{\ensuremath{\mathsf{Claim}_{F}}}
\newcommand{\pr}{\ensuremath{\mathsf{Pr}}}
\newcommand{\op}{\ensuremath{\mathsf{Def}}}
\newcommand{\com}{\ensuremath{\mathsf{Com}}}

\newcommand{\To}{\Rightarrow}

\newcommand{\non}{\ensuremath{\mathcal{\sim }}}

\newcommand{\set}[2][\relax]{\ensuremath{#1\{#2#1\}}}
\newcommand{\Set}[2][\relax]{%
  \ensuremath{
    \ifx#1\left
      #1\{#2\right\}
    \else\ifx#1\right
      \left\{#2#1\}
      \else #1\{#2#1\}\fi\fi}%
}

\newcommand{\OBL}{\Obl}

\def\dRule#1:#2=>#3#4{#1\colon #2\To_{#3}#4}

\setcopyright{rightsretained}
\copyrightyear{2025}
\acmYear{2025}
\acmDOI{XXXXXXX.XXXXXXX}
\acmConference[ICAIL 2025]{International Conference on Artificial Intelligence and Law}{June 16--20,
  2025}{Chicago, IL}
\acmISBN{978-1-4503-XXXX-X/18/06}

\begin{document}

\title{Judicial Permission}

\author{Guido Governatori}
%\authornotemark[1]
\orcid{0000-0002-9878-2762}
\affiliation{%
  \institution{Central Queensland University}
  \city{Rockhampton}
  \state{Queensland}
  \country{Australia}
}
\email{g.governatori@cqu.edu.au}
 
\author{Antonino Rotolo}
\affiliation{%
  \institution{University of Bologna}
  \city{Bologna}
  \country{Italy}}
\email{antonino.rotolo@unibo.it}

\renewcommand{\shortauthors}{Governatori and Rotolo}

\begin{abstract}
This paper examines the significance of weak permissions in criminal trials (\emph{judicial permission}). It introduces a dialogue game model to systematically address judicial permissions, considering different standards of proof and argumentation semantics.

\end{abstract}




\maketitle

\section{Introduction}
\label{sec:interoduction}

The notion of permission is one of the fundamental concepts in deontic logic and legal reasoning.  Permission is a multi-faceted concepts with many different aspects.  A general definition of permission is the absence of prohibition or the lack of the obligation to the contrary \cite{alchourron-bulygin:1984,alchourron1971normative}. Multiple variants of permission have been identified in the literature, such as \emph{Strong Permission} and \emph{Weak Permission} (see \cite{handbook:permission,MakinsonT03}.  

Strong permission is often interpreted as the presence of a norm making something explicitly permitted, and in general this norm derogates a prohibition. On the contrary, it is possible to identify two main cases to establish weak permission: the absence of norm regulating a certain behaviour or the norms prescribing a certain behaviour are not applicable in a particular context.

Here we will focus on the second case, where the norms prescribing a certain behaviour are not applicable in a particular context.  In particular, we focus on the situation where the decision that some norms establishing obligations or prohibitions are deemed not applicable as a result of a court decision. We call the resulting weak permissions \emph{Juridical Permissions}.

When we consider a court decision, in particular for a penal case, we have to interface reasoning with the appropriate statutes, and the evidence of the case. Moreover, the two forms of reasoning (reasoning with statutory norms and reasoning with evidence) might be subject to different standards of proof.

In this paper, we will present a formal model of \emph{Judicial Permission} based on the framework of \emph{Defeasible Deontic Logic} \cite{jpl:permission}. Defeasible Deontic Logic is a formal framework for representing and reasoning about norms and permissions that allows for the modelling of judicial permission. We will show how judicial permission can be represented in defeasible deontic logic, and how it can be used to reason about them in legal reasoning.

In criminal trials,  judicial/weak permission gains a critical operational significance (see \cite{GovernatoriR23}. Unlike strong permissions, which are explicitly codified, judicial/weak permissions arise from the absence of explicit prohibitions. They are elusive because they are inferred from the legal system’s allowance or lack of prohibition for certain actions.
Judges often determine whether an action is weakly permitted, but this is complicated by the absence of direct provisions in criminal codes or case law precedents \cite{criminal}. This indeterminacy is exacerbated in adversarial legal systems, where parties have different sets of evidence and rules, shaping their strategies.

This paper conceptualises judicial permissions in criminal procedures within dialogue games. These games, characterised by argumentative exchanges, provide a structured model where prosecution and defence navigate through a blend of common and private knowledge, striving to establish or negate claims.

By employing different argumentation semantics corresponding to diverse standards of proof, we aim to unravel the subtleties of weak permissions from a procedural perspective. A detailed exploration will illuminate the procedural underpinnings through which judicial permissions are construed and managed, providing valuable insights into the adjudicative processes underpinning criminal law.

The layout of the paper is as follows. Section \ref{sec:dialogues1} offers some intuitions for understanding weak permissions in criminal proceedings. Section \ref{sec:standards} establishes a conceptual connection between standards of proofs criminal proceedings and  judicial permission. Section \ref{sec:dialogues2} provides a formal definition of criminal-proceeding dialogue games where judicial permission may be obtained while Section \ref{sec:weak_dialogues} characterizes the notion of judicial permission in criminal-proceeding dialogue games.  

\section{Deontic dialogues in criminal proceedings: intuition}\label{sec:dialogues1}

In this section we pave the way for developing a framework where the concept of judicial permission takes place in criminal trials. The idea is that analyzing \textit{judicial permission} within the dynamics of a criminal trial is founded on the understanding that it is the judge's responsibility to determine whether an action is weakly permitted \cite{criminal}. This responsibility arises precisely because what is deemed weakly permitted is not clearly established by the criminal code nor by any precedent in criminal case law. If it were otherwise, the action would be explicitly permitted, and thus not weakly permitted. Accordingly, it is only through the procedural dynamics that we are able to provide a more precise account of the notion of judicial permission in criminal law. The absence of well-defined legal precedents makes the judge's role critical in categorizing actions that are not plainly permitted.

In \cite{GovernatoriORSS14} it was argued that court proceedings exemplify the so-called argument games with \emph{incomplete information}, i.e.,  dialogues where the structure of the game is \emph{not}
common knowledge among the players. The concept has been later formally and computationally investigated \cite{ecai2014,GovernatoriMORS14} and extended  to also model legislative dialogues \cite{GovernatoriR19,GovernatoriRRV19}. 

In criminal proceedings, dialogues often involve incomplete information as the prosecution and defendant are unaware of their opponent’s arguments. These arguments can be modelled by assuming players have different private knowledge theories that are unknown to their opponents. Players can build arguments using their private knowledge, while opponents construct new arguments by reusing disclosed information and their private knowledge to counter the opponent’s arguments.

In criminal cases, the prosecution ($\pr$) initiates a dialogue ($\mathcal{D}$) to prove the defendant ($\op$)’s guilt beyond a reasonable doubt. The dialogue game involves alternating interactions between $\pr$ and $\op$. $\pr$ assesses the validity of a claim, while $\op$ has the burden of proof on the opposite claim. The challenge is formalised by argument exchange. Players share a logical theory with common facts and rules, but each has private knowledge of some rules. Other rules are known by both, but the set may be empty. These rules and facts represent common knowledge. Each turn involves either putting forward an argument or passing. By putting forward a private argument, the agent increases common knowledge by the rules used. Essentially, they choose a subset of private knowledge that justifies their claim. When a player passes, they declare defeat and the game ends if no combination of remaining private rules proves their claim.

$\pr$'s  initial claim ($\claim$) typically consists in two sets of statements: a set 
\[\fclaim=\set{a_1, \dots , a_n}\]
of \emph{evidential claims} (such as ``$\op$ did $b$'') and a set 
\[\dclaim=\set{\OBL \non a_1, \dots , \OBL \non a_n}\] 
of \emph{deontic claims} (such as ``$b$ is prohibited by the criminal code''). In a similar way, we  have two different types of rules: \emph{evidential rules}, having the form 
\[
r: a_1, \dots , a_n \To b
\]
and \emph{deontic rules} having the form 
\[
s: a_1, \dots , a_n \To_{\OBL} b
\]
if $r$ is applicable then we can derive that $b$ is the case, if $s$ is applicable then we can prove that $b$ is obligatory. 
%The different logical theories typically consist of diverse sets of rules.
The set $R$ of all evidential and deontic rules, which are used to build arguments, is partitioned into three subsets: a set
$R_{\com}$ known by both players and two subsets $R_{\pr}$ and $R_{\op}$
corresponding, respectively, to $\pr$'s and $\op$'s private knowledge. 

Consider a setting where $F=\set{a,d,f,g}$ is the common knowledge of indisputable facts, $R_{\com}=\emptyset$, and the players have in $\mathcal{D}$ the following rules:
\begin{align*}
%    F  & = \{  a, d, f\} \\
%    R_{\com} & =  \emptyset \\
    R_{\pr} = \{& r_1\colon a\To b ,\;  
                     r_2\colon d\To c ,\;
                     r_3\colon c \To b, 
                     r_4\colon g \To_{\OBL} \non b\}\\
    R_{\op} = \{& r_4\colon c\To e ,\;
                       r_5\colon e,f\To \neg b, r_6\colon c \To_{\OBL} \non b\}.
\end{align*}

If $\pr$'s claim is 
\[
    \claim_{\pr} = \langle \fclaim = \set{b}, \dclaim = \set{\OBL \non b}\rangle
\]
and $\Pr$ plays 
\[
\set{r_1\colon a\To b,\; r_4\colon g \To_{\OBL} \non b},
\]
then $\pr$ wins the game. If $\pr$ plays 
\[
\set{r_2\colon d\To c,\; r_3\colon c\To b, \; r_4\colon g \To_{\OBL} \non b}
\]
(or even $R_{\pr}$), this allows $\op$ to succeed. Here, a minimal subset of $R_{\pr}$ is
successful. 

The situation above can be reversed for
$\pr$. Replace the sets of private rules with
\begin{align*}
%    F  & = \{  a, d, f\} \\
%    R_{\com} & =  \emptyset \\
    R_{\pr} = \{& r_1\colon a\To b ,\;  
                     r_7\colon d\To \non c ,\;
                     r_4\colon g \To_{\OBL} \non b\}\\
    R_{\op} = \{& r_7\colon f\To c ,\;
                       r_8\colon d,c\To \neg b, r_6\colon c \To_{\OBL} \non b\}.
\end{align*}
Playing 
\[
\set{r_1\colon a\To b,\; r_4\colon g \To_{\OBL} \non b}
\]
is now not successful for $\pr$, while the whole $R_{\pr}$ ensures victory.

What lesson can we learn from the above discussion? It seems to us that the following intuition holds.

\begin{intuition}\label{int:basic}[judicial permissions in dialogues]
Let $\mathcal{D}$ be a dialogue where $\pr$'s  claim against $\op$ consists in the sets of statements 
\[\fclaim=\set{a_1, \dots , a_n}
\qquad\qquad
\dclaim=\set{\OBL \non a_1, \dots , \OBL \non a_n}.
\]
If $\pr$ succeeds in $\mathcal{D}$, each $a_k$, $1\leq k \leq n$, is not weakly permitted, while if $\op$ succeeds in $\mathcal{D}$  each $a_k$ is weakly permitted.
\end{intuition}

The above discussion provides an initial understanding of when judicial permissions can be obtained in dialogues modelling criminal trials. To account for this, we can adjust the strategic dialogue idea from \cite{ecai2014,GovernatoriMORS14} to cover interactions between evidential and deontic claims. However, different proof standards can be used, which may reopen the question of whether semantics like grounded semantics can logically characterise judicial permission.

\section{Proof standards and judicial permissions}\label{sec:standards}
Intuition \ref{int:basic} is based on a dialogue setting where we do not pay attention to the role standards of proof. However, in common-law systems, procedural law (especially in civil cases) is well known for distinguishing between various legal standards.\footnote{On the other hand, civil-law systems employ a standard for civil cases that closely mirrors the standard used in criminal cases: the judge must be firmly convinced of the truth of the alleged facts.} These standards vary by case type, with more serious consequences typically requiring a higher standard. 

Common evidential proof standards, which have been notably studied in the AI\&Law community since \cite{GordonWalton:proof, BrewkaGordon:comma10}, are:
\begin{itemize}
  \item \emph{scintilla of evidence},
  \item \emph{preponderance of evidence}, also called \emph{best argument} in \cite{GordonPrakkenWalton:Carneades},
  \item \emph{clear and convincing evidence},
  \item \emph{beyond reasonable doubt},
  \item \emph{dialectical validity}
\end{itemize}
The above proof standards are listed in order of strength, this means that the
conditions to satisfy them are more and more stringent. In addition a stronger
proof standard includes the weaker ones. 

Governatori has reconstructed these standards in the context of Defeasible Logic (DL) \cite{icail2011carneades}.  In DL proof theory is based on the concept of a \emph{defeasible theory}, which is a  structure $D=(F,R,>)$ where $F$ is a set of facts, $R$ is a set of rules, and $>$, is the superiority relation over the set of rules that allows for handling conflicting arguments. On this bases, we can distinguish several \emph{types of conclusions}. A \emph{conclusion} of $D$ is a
tagged literal and can have one of the following forms:
\begin{enumerate}
  \item $+\partial l$: $l$ is defeasibly provable in $D$ using the ambiguity blocking variant of DL (a skeptical proof standard corresponding to standard argumentation semantics for DL);
  \item $-\partial l$: $l$ is defeasibly rejected in $D$ using the ambiguity
  blocking variant;
  \item $+\delta l$: $l$ is defeasibly provable in $D$ using the ambiguity
  propagation variant of DL (a proof standard corresponding to grounded semantics);
  \item $-\delta l$: $l$ is defeasibly rejected in $D$ using the ambiguity
  propagation variant;
  \item $+\sigma l$: $l$ is supported in $D$, i.e., there is a chain of
  reasoning leading to $l$ (in other words, we check for a forward chaining of rules to
propagate the `support' for rules, but  do not propagate the
support from the premises to the conclusion in case we have a rule for the
contrary unless the rule is not weaker than a rule whose premises are all
provable);
  \item $-\sigma l$: $l$ is not supported in $D$;  
   \item $+\sigma^- l$: $l$ is weakly supported in $D$, i.e., there is a chain of
  reasoning leading to $l$ (we simply check for a forward chaining of rules to
propagate the `support' for rules, with no constraints);
  \item $-\sigma l$: $l$ is not weakly supported in $D$.  
\end{enumerate}
The proof tags determine the strength of a derivation. The proof tags $+\delta$,
$-\delta$, $+\partial$ and $-\partial$ are for skeptical conclusions, and
$+\sigma$, $-\sigma$, $+\sigma^-$ and $-\sigma^-$ capture credulous conclusions.

\begin{proposition}\label{pro:def:inclusion} \cite{tocl:inclusion,icail2011carneades}
  For any theory $D$ and literal $l$, the following implications hold:
  \setlength{\columnsep}{3pt}
  \vspace*{-10pt}
  \begin{multicols}{2}
  \begin{enumerate}[wide]
    \item $D\vdash+\delta l$ implies $D\vdash+\partial l$;
    \item $D\vdash+\partial l$ implies $D\vdash+\sigma l$;
    \item $D\vdash+\sigma l$ implies $D\vdash+\sigma^- l$;
    \item $D\vdash-\partial l$ implies $D\vdash-\delta l$.
    \item $D\vdash-\sigma l$ implies $D\vdash-\partial l$;
    \item $D\vdash-\sigma^- l$ implies $D\vdash-\sigma l$;
  \end{enumerate}
  \end{multicols}
\end{proposition}

Proposition \ref{pro:def:inclusion} allows for establishing the relative strength of proof standards:

\begin{definition}\label{def:strenght}[Relative strength of proof standards]
The relative strenght of proof standards is represented as follows:
\begin{gather*}
+\delta \prec +\partial \prec +\sigma \prec +\sigma^- \qquad\qquad
-\sigma^-  \prec -\sigma \prec -\partial \prec -\delta.
\end{gather*}

\end{definition}

It was argued in \cite{icail2011carneades} that, for any defeasible theory $D=(F,R,>)$,
\begin{itemize}
    \item $l$ is proved in $D$ with proof standard scintilla of evidence iff $D \vdash +\sigma^- l$,
    \item $l$ is proved in $D$ with proof standard substantial evidence iff $D \vdash +\sigma l$,
    \item $l$ is proved in $D$ with proof standard preponderance of evidence iff $D \vdash +\partial l$,
    \item $l$ is proved in $D$ with proof standard beyond reasonable doubt iff $D \vdash +\delta l$,
     \item $l$ is proved in $D$ with proof standard beyond reasonable doubt iff $D \vdash +\delta l$,
     \item $l$ is proved in $D$ with proof standard dialectical validity iff $D' \vdash +\delta l$, where $D' = (F, R, \emptyset)$.
 \end{itemize}
Dialogue-based approaches like those in Section \ref{sec:dialogues1} are insufficient to prove the prosecution’s claim. The claim is based on two elements: evidential claims focused on evidence issues and deontic claims focused on deontic statements. Defeasible Deontic Logic (DDL) can help \cite{handbook:deontic} because it extends DL to cover deontic rules and operators, inheriting proof theory features to accommodate deontics. DDL is standard DL plus deontic operators $\Obl$ and $\Perm$ for obligations and permissions. It also uses two types of rules: standard (non-deontic, evidential) rules with arrow $\To$ and deontic rules.  
\[
r\colon \seq{a} \To_{\Obl} b.
\]
If $r$ is applicable (i.e., $\seq{a}$ are proved), then we derive $\Obl b$. This formalism (similar to other logics like \cite{MakinsonT03}) allows identifying various reasoning patterns for deriving permissions, including judicial permissions. A permission $\Perm a$ is a judicial permission if $\Obl \neg a$ is not provable (i.e., $\neg \Obl \neg$ implies $\Perm$). Since we have two consequence relations, we duplicate proof tags and conclusion types. For instance, $+\partial_{\OBL} l$ means $l$ that is defeasibly provable (using ambiguity blocking) with mode $\OBL$, i.e., as an obligation. Similarly, $-\partial_{\OBL} \neg l$ allows proving that $l$ is weakly permitted. $+\delta l$ means $l$ that is defeasibly provable (using ambiguity propagation) as an evidential conclusion.

We emphasised in \cite{GovernatoriR23} that judicial permission in criminal law plays a peculiar role in relation to the principle of legality, which ensures the deontic closure of the criminal legal system. The closure rule \emph{nullum crimen sine lege} states that any action not prohibited is permitted by criminal law. However, when a judicial permission $X$ is used to derive deontic conclusions that are not in favour of the defendant, it is an \emph{afflictive judicial permission} and may not be derived on the basis of the principle of legality. We extended DDL into DDL$^+$ to explicitly state all proof standards required by criminal procedure to propagate deontic ambiguity in reasoning chains. DDL$^+$ adjusts the language and proof theory of DDL to handle rules such as 
\[
    r\colon +\delta a, +\partial c, -\delta\Obl\neg c \Rightarrow_\Obl d.
\]
The intuitive meaning of the rule is that: we infer that $d$ is obligatory provided that $a$ must be positively proved using the ambiguity propagation standard, $c$ must be positively proved using the ambiguity blocking standard, and $\Obl\neg c$ must be refuted with the ambiguity propagation standard.

\section{Deontic dialogues in criminal proceedings: formal definition}\label{sec:dialogues2}

We are now in the condition to provide the definition of a criminal-proceeding dialogue game that takes into account proof standards. The dialogue is nothing but a sequence of moves initiated by $\pr$ with a criminal charge called claim and where $\pr$ and $\op$ exchange arguments. The state of the game at any turn $i$ is denoted by a defeasible theory $D^i = (R^i_{\com}, >)$ and by two sets $R^i_{\pr}$ and $R^i_{\op}$. $R^i_{\com}$ is the set of rules known by both
$\pr$ and $\op$ at turn $i$ (which may be empty when $i=0$), and $R^i_{\pr}$
($R^i_{\op}$) is the private knowledge of $\pr$ ($\op$) at turn $i$. We assume
that each party is informed about the restriction of superiority relation $>$ to the rules that she
knows. We assume that the private theories of $\pr$ and $\op$ are conflict-free.

We now discuss the game rules which establish how the theory $D^{i-1}$ and the sets $R^{i-1}_{\pr}$, $R^{i-1}_{\op}$ are modified based on the move played at turn $i$.

The prosecution starts the game by choosing the content of dispute $\claim = \langle \fclaim, \dclaim \rangle$. At turn $i$, $\pr$ has the burden to justify elements in $\fclaim$ and $\dclaim$ by using the current common knowledge along with a subset of $R_{\pr}^{i-1}$, whereas $\op$ must refute elements in $\fclaim$ and $\dclaim$. We point out that at turn $i$, $\pr$ ($\op$) may put forward arguments with conclusions different from those proving or refuting elements in $\fclaim$ and $\dclaim$.

Criminal-proceeding dialogue games are based on the following assumptions:
\begin{itemize}
  \item the burden of proof\footnote{AI\&Law research has discussed different types of burden (mainly focused on civil cases) \cite{GordonWalton:proof}. 
  In this paper, we do not consider all the types, but we generically speak of burden of proof, even though we mainly focus on the so called burden of persuasion.} falls on the prosecution $\pr$\footnote{There are exceptions. For example, in criminal cases where a defense of insanity is raised, it is the responsibility of the defense to establish this claim on a balance of probabilities.};
  \item for the evidential claims in $\fclaim$ the the proof standard is beyond reasonable doubt\footnote{There are also here exceptions. Again, where a defense of insanity is raised, the defense must establish this claim adopting standards of civil cases.};
  \item for the deontic claims in $\dclaim$, any specification of standards is directly stated in deontic rules, since such requirements are usually explicitly codified in criminal-procedure provisions.
\end{itemize}

\begin{definition}[Criminal-proceeding dialogue game]\label{def:free_dialogue}
Let $R^{0}_{\com}$, $R^{0}_{\pr}$ and $R^{0}_{\op}$ be respectively the common knowledge, and the private knowledge of $\pr$ and $\op$ at the beginning of the dialogue game $\mathcal{D}$, and $D^0=(R^0_{\com}, >)$. 
As notational convention, we write  $\#\in \set{\delta, \partial, \sigma, \sigma^-}$.

A \emph{Criminal-proceeding dialogue game} $\mathcal{D}$ is a finite sequence of moves defined as follows:

\begin{itemize}
 \item At turn $0$, $\pr$ starts the game if, for some sets 
 \[\fclaim=\set{a_1, \dots , a_n}
 \qquad\qquad
 \dclaim=\set{\OBL \non a_1, \dots , \OBL \non a_n},\] 
 for each $a_k$, $1 \leq k \leq n$, we have that $D^0\vdash +\delta a_k$ and $D^0\vdash +\#_{\OBL} \non a_k$.

\item At turn $i$, $i>0$, if $\pr$ plays $R^{i}$, then, there is a set of literals $\set{l_1, \dots , l_k}$ such that, for each $l_j$, $1 \leq j \leq k$,
\begin{itemize}
    \item
    \begin{enumerate*}
    \item[(i)] $D^{i-1}\vdash +\# l_j$, or 
    \item[(ii)] $D^{i-1}\vdash -\# l_j$, or 
    \item[(iii)] $D^{i-1}\vdash +\#_{\OBL} l_j$, or 
    \item[(iv)] $D^{i-1}\vdash -\#_{\OBL} l_j$;
    \end{enumerate*}
    \item $R^{i}\subseteq R^{i-1}_{\pr}$;
	\item $D^{i} = (R^{i}_{\com},>)$;
	\item $R^{i}_{\pr} = R^{i-1}_{\pr}\setminus R^{i}$, $R^{i}_{\op} = R^{i-1}_{\op}$, and $R^{i}_{\com} = R^{i-1}_{\com}\cup R^{i}$;
    \item 
    \begin{enumerate*}
	\item[(i)] $D^{i}\vdash +\# \non l_j$ or $D^{i}\vdash -\# \non l_j$, or $D^{i}\vdash +\delta \non l_j$ if $l_j\in\fclaim$, or
    \item[(ii)] $D^{i}\vdash +\# l_j$, or $D^{i}\vdash +\delta l_j$ if $l_j\in\fclaim$, or\\ 
    \item[(iii)] $D^{i}\vdash +\#_{\OBL} \non l_j$ or $D^{i-1}\vdash -\#_{\OBL} l_j$, or 
    \item[(iv)] $D^{i}\vdash +\#_{\OBL} l_j$;
    \end{enumerate*}
\end{itemize}

\item At turn $i$, $i>0$, if $\op$ plays $R^{i}$, then, there is a set of literals $\set{l_1, \dots , l_k}$ such that, for each $l_j$, $1 \leq j \leq k$,
\begin{itemize}
	\item 
    \begin{enumerate*}
    \item[(i)] $D^{i-1}\vdash +\# l_j$, or 
    \item[(ii)] $D^{i-1}\vdash -\# l_j$, or 
    \item[(iii)] $D^{i-1}\vdash +\#_{\OBL} l_j$, or (iv) $D^{i-1}\vdash -\#_{\OBL} l_j$;
    \end{enumerate*}
	\item $R^{i}\subseteq R^{i-1}_{\op}$;
	\item $D^{i} = (R^{i}_{\com},>)$;
	\item $R^{i}_{\op} = R^{i-1}_{\op}\setminus R^{i}$, $R^{i}_{\pr} = R^{i-1}_{\pr}$, and $R^{i}_{\com} = R^{i-1}_{\com}\cup R^{i}$;
    \item \begin{enumerate*}
	\item[(i)] $D^{i}\vdash +\# \non l_j$ or $D^{i}\vdash -\# \non l_j$, or 
    \item[(ii)] $D^{i}\vdash +\# l_j$, or \\
    \item[(iii)] $D^{i}\vdash +\#_{\OBL} \non l_j$ or $D^{i-1}\vdash -\#_{\OBL} l_j$, or (iv) $D^{i}\vdash +\#_{\OBL} l_j$;
    \end{enumerate*}
\end{itemize}
 \item At turn $i$ $\mathcal{D}$ terminates if
  \begin{itemize}
   \item {\bf $\op$ succeeds}: $R^{i-1}_{\pr} = \emptyset$ and there is any $a_k$, $1\leq k \leq n$, such that 
   \begin{enumerate}
        \item[(a)] $D^{i-1}\vdash -\#_{\OBL} \non a_k$ or $D^{i-1}\vdash +\#_{\OBL} a_k$, or 
        \item[(b)]  $D^{i-1}\vdash +\# \non a_k$ or $D^{i-1}\vdash -\# a_k$;
   \end{enumerate}
   \item {\bf $\pr$ succeeds}: $R^{i-1}_{\op} = \emptyset$ and for each $a_k$, $1\leq k \leq n$, $D^{i-1}\vdash +\delta a_k$ and $D^{i-1}\vdash +\#_{\OBL} \non a_k$.
  \end{itemize} 
 \end{itemize}
 \end{definition}
 
As we informally illustrated in Section \ref{sec:dialogues1}, at each turn $i$, we update the theory $D^{i-1}$ by adding to $R^{i-1}_{\com}$ the private rules played at turn $i-1$ by the party at stake in response to the rules played by the opponent. We should note that in Definition \ref{def:free_dialogue} the rule update is free: the revision of $R^i_{\com}$ is not constrained by anything and we simply check the capability of $\pr$ of proving the claims and of $\op$ of refuting them. On the contrary, although this can be theoretically possible, it is well known that the critical steps must be usually approved or decided by judges or by the jury. For the sake of simplicity, we omit this aspect, which does not affect the nature of judicial permissions.\footnote{To capture this aspect, following the model proposed in \cite{GovernatoriR19} for legislation, we could introduce a function $\ochoice$ expressing the decision of the judge or the jury for each turn $i$, and which maps a defeasible theory $D^i$ (which is what results from adding the rules played by parties at $i-1$) into a possibly different defeasible theory $D'^i$: the resulting theory---and the corresponding conclusions---is what the judges or the jury approve in the trial for the turn at stake.}

\section{Judicial permission in criminal-proceeding dialogue games}\label{sec:weak_dialogues}

This section draws logical conclusions about how the dialogue game framework can determine if an action or statement is weakly permitted in criminal proceedings. As defined in Definition \ref{def:free_dialogue}, the dialogue mechanism allows $\pr$ and $\op$ to exchange arguments systematically, making it suitable for exploring judicial permissions. This investigation relies on DDL, which assesses permission status based on dialogue game rules and evidence.

Note that the following result for DDL can be obtained by extending Proposition 3 of \cite{jpl:permission} using techniques from \cite{tocl:inclusion}. This result provides a foundational logic for understanding weak permissions in this context.
\begin{proposition}
  \label{prop:oblperm}
  Let $D=(F, R,>)$ be a Defeasible Theory such that $>$ is acyclic and for any literal $l$, $F$ does not contain any pairs like $\OBL l$ and $\OBL\non l$. For any literal $l$ and $\#\in \set{\delta, \partial}$, if $D\vdash+\#_{\OBL} l$, then $D\vdash-\#_{\OBL} \non l$.
\end{proposition}
Proposition~\ref{prop:oblperm}  establishes a fundamental distinction in how sceptical and credulous standards affect the interpretation of permissions. Specifically, it underscores the impossibility of simultaneously having proof for both $\OBL l$ and $\OBL \neg l$ in sceptical frameworks—a key observation ensuring consistency in legal adjudications relying on deontic logic.

However, proposition does not hold for $\#\in\set{\sigma, \sigma^-}$, which derives from a basic property of $\#\in\set{\delta, \partial}$ only. Notably, for a theory $D$ satisfying the conditions of Proposition \ref{prop:oblperm}, and for any literal $l$, it is not possible to have both $D \vdash +\#_{\OBL} l$ and $D \vdash +\#_{\OBL} \non l$ \cite[Proposition 2]{jpl:permission}. On the contrary, these conflicting conclusions may both obtain with $\sigma$ and $\sigma^-$, because these are credulous standards, hence they allow for arguments supporting both $\OBL l$ and $\OBL \non l$, reflecting hard difficulties to describe judicial permissions in credulous deontic reasoning and in more permissive logical scenarios often found in complex legal disputes.

The broader implication here echoes in the following result, typical for judicial permissions, which builds on Proposition \ref{prop:oblperm} and highlights its relevance primarily with sceptical standards \cite[Proposition 4]{jpl:permission}.

\begin{proposition}\label{th:ought-can}
 Let $D=(F, R,>)$ be a Defeasible Theory such that $>$ is acyclic and for any literal $l$, $F$ does not contain any pairs like $\OBL l$ and $\OBL\non l$. For any literal $l$ and $\#\in \set{\delta, \partial}$, if \mbox{$D\vdash +\#_{\OBL} l$}, then $l$ is weakly permitted.
\end{proposition}
Notice that Proposition~\ref{th:ought-can} requires a theory to be (deontically) consistent. Thus, if the theory proves $+\#_\Obl l$, then this ensures that the theory does not prove $+\#_\Obl\non l$; hence, the theory fails to obtain $\non l$ as an obligation, and thus we can establish $l$ as a judicial permission. Accordingly, we can formulate the following definition for when a literal is weakly permitted according to a theory. 

\begin{definition}
\label{def:weak_permission}
Let $D$ be a Defeasible Theory and $\#\in \set{\delta, \partial}$. A literal $l$ is $\#$-\emph{weakly permitted} iff 
$D\vdash -\#_{\OBL} \non l$.
\end{definition}

Definition \ref{def:weak_permission} may account for the concept of judicial permission with respect to any specific theory $D^i$ at each turn of $i$ of a criminal-proceeding dialogue game $\mathcal{D}$. However, the following definition clearly follows from Definition \ref{def:weak_permission}:

\begin{definition}
\label{def:D-weak_permission}
Let $\mathcal{D}$ be any criminal-proceeding dialogue game where $\dclaim =\set{\OBL \non l_1, \dots , \OBL \non l_n}$ and $\#\in \set{\delta, \partial}$. Any literal $l_k$, $1 \leq k \leq n$, is $\#$-\emph{weakly permitted} in $\mathcal{D}$ iff
\begin{itemize}
 \item $\mathcal{D}$ terminates at turn $i$;
 \item $D^{i-1}\vdash -\#_{\OBL} \non l_k$.
\end{itemize}
\end{definition}

These definitions reflect how the abstract concept of judicial permission is applied within the operational context of a criminal-process dialogue game, providing the logical basis upon which legal permissions are inferred absent explicit prohibitions.

\balance

\section{Conclusion}\label{sec:conclusion}

In this paper, we explore the significance and complexities of weak permissions in criminal trials. We challenge the notion that weak permissions simply equate to duals of obligations, arguing for their greater depth and relevance in legal frameworks. Our research reveals that weak permissions determine the boundaries of lawful behaviour in criminal law, especially in adversarial systems with incomplete information.

We introduced a structured dialogue game model to capture the dynamics between prosecution and defence, showing how judicial permissions can be addressed through exchanges of common and private knowledge. By using various argumentation semantics, we offer a nuanced understanding of judicial permissions, providing insights into their procedural implications in criminal adjudication.

Our study contributes to the broader discourse on permissive norms, advancing towards a more structured approach in formal legal judgements. It emphasises the need to accommodate judicial permissions in legal reasoning, highlighting their implicit yet influential role in judicial decisions.

The following are directions for future work.

\begin{itemize}

    \item \textbf{Cross-Legal-System Analysis:} A promising direction is the comparative analysis of judicial permissions across different legal systems—common law versus civil law—to evaluate how varying legal traditions influence the treatment and interpretation of permissive norms.
    
    \item \textbf{Logical Investigation of Dialogue Games:} The criminal-proceeding dialogue game is based on the basic strategic argument game with incomplete information as discussed in \cite{ecai2014,GovernatoriMORS14}, which has primarily been studied from a computational viewpoint. An intriguing line of future work is to develop a systematic formal investigation of the logical and deontic properties characterizing these types of games, especially when various types of normative and deontic statements are involved.
\end{itemize}

\printbibliography

\end{document}